\documentclass[jou,apacite]{apa6}
\title{CAMREP- Concordia Action and Motion Repository}
\usepackage{authblk}
\usepackage[T1,hyphens]{url}

\author{Kaustubha~Mendhurwar,
        Qing~Gu,
        Vladimir de la Cruz,
        Sudhir~Mudur,
        and~Tiberiu~Popa}
\affiliation{Concordia University, Montreal, Quebec, Canada 
}

\abstract{
Action recognition, motion classification, gait analysis and synthesis are fundamental problems in a number of fields such as computer graphics, bio-mechanics and human computer interaction that generate a large body of research. This type of data is complex because it is inherently multidimensional and has multiple modalities such as video, motion capture data, accelerometer data, etc.
While some of this data, such as monocular video are easy to acquire, others are much more difficult and expensive such as motion capture data or multi-view video. This creates a large barrier of entry in the research community for data driven research. We have embarked on creating a new large repository of motion and action data (CAMREP) consisting of several motion and action databases. What makes this database unique is that we use a variety of modalities, enabling multi-modal analysis.  Presently, the size of datasets varies with some having a large number of subjects while others having smaller numbers. We have also acquired long capture sequences in a number of cases, making some datasets rather large.
}

\begin{document}
\maketitle    
  
\section{Introduction}

Standardized datasets for specific applications are an invaluable resource for the academic community providing critical bench-marking tools for algorithm design and more objectivity through quantitative evaluation of newly proposed algorithms and techniques. 
Time series data are ubiquitous with motion and gesture data becoming more and more popular due to emerging technologies such as the Kinect devices~\cite{Kin} or more accurate accelerometer data~\cite{Myo}. However, there is a dearth of large, high-dimensional and multi-modal time series data repositories. 
Therefore, in our research in time series matching~\cite{Kaus} we have created a repository consisting of several databases, using a variety of modalities. One of the datasets is much larger as compared to others currently available publicly in the same category. It constitutes data of $97$ subjects and with at least two minutes of activity by each subject. 
For the benefit of further research using such datasets, we have made this large dataset publicly available for download at the following web address:\url{https://users.encs.concordia.ca/~graphics/camrep/}. If you use this database please cite~\cite{Kaus}, which was the original research that produced this data. 

\section{Comparison with Other Publicly Available Databases}

The UCR database~\cite{UCRArchive} is a large time series database that consists of 85 different datasets coming from many different domains. It is commonly used for time series matching and it is constantly growing. At present, most signals in the dataset are low dimensional signals (most often single dimension).   

The \textit{MSRC-12} is a gesture dataset from Microsoft Cambridge~\cite{msrc12} created using the Microsoft Kinect v2 device that features higher dimensional data. The \textit{MSRC-12} Kinect gesture data set consists of
motion data for $12$ relatively similar gestures in up to $6,244$ annotated action sequences.
The domain of this signal is the Kinect pose estimation system, which provides the position of $20$ markers in Euclidian space.

For motion database, one of the most commonly used databases is the CMU database~\cite{CMU}. It contains a large collection of motions captured using a VICON motion capture system. The sequences in this database are however relatively short in length.

Our repository of gait and action data consists of several datasets produced using a variety of modalities, which include Kinect, Myo device and Vicon motion capture system. Our gait datasets are two minutes long for each subject. The captured database consists of subjects ranging over different age, sex, physique, ethnicity, etc. More details of the content and the capture process follow.
                        
\section{Concordia Action and Motion Repository}

\subsection{VICON Motion Capture Datasets}
The first database we present is a gait database. 
We used a VICON motion capture system available at the PERFORM center in Montreal and we created a database of $97$ subjects of varying characteristics. 
For compattibility with available data in this category, we used the same configuration of markers and angles as the CMU database~\cite{CMU}.
Since the capture volume of the available system is relatively small, limiting the recording to just $1-2$ walk cycles at a time, we placed a treadmill in the scene so as to be able to capture longer sequences.
We had to remove the sides of the treadmill in order to avoid occlusions during capture.
Studies in Kinematics have shown that walking on treadmill or overground are very similar and the magnitude of the differences was comparable to the normal variability of gait parameters~\cite{Riley200717}.
Therefore, our acquisition of gait data on the treadmill is not significantly different to the acquisition of overground gait.

We captured at least two separate sessions of one minute each for every one of the 97 subjects. A one minute walk has approximately $50$ walking cycles. Data capture was done over a period of 45 days. Subjects from public were invited (through a printed poster) to participate in this experiment and were awarded a small gift. 

From this data capture exercise, we provide the following datasets:
\begin{enumerate}
\item (A) walking cycle data in kinematic angle space
\item (B) walking cycle data in kinematic joint space (hole filled)
\item (C) walking cycle data in kinematic joint space (raw data with holes)
\end{enumerate}

Due to occlusions, the marker paths typically have missing segments. The state of the art VICON capture pipeline has a semi-automatic method to fill these gaps, but it is still a tedious process that took a technician $2-3$ days. Dataset (B) is the data with technician filled in data. Dataset (C) contains the raw data, where we put a $0$ for the values that are missing. 
For all these datasets we also extract the walk cycles using a right foot down convention providing the start of each walk cycle.

\subsection{Kinect Motion Capture Datasets}
Using a similar treadmill setup as described earlier for the Vicon system, we created some what smaller walking datasets for similar activity using the Kinect device: Dataset (D) and Dataset (E). 
In contrast to the VICON data, the Kinect data is less reliable and more noisy.
The Kinect pose estimation system provides the position of $20$ markers.
Since our application domain is gaits, we consider the $9$ markers corresponding to the lower body.
Dataset (D) has walking data for $21$ subjects; two one minute sequences, both acquired using flat settings on the treadmill.
Dataset (E) has data for $10$ subjects walking on different inclines. 
We captured two half minute sequences for every different incline setting on the treadmill from $0$ degrees to the maximum slope of the treadmill of $25$ degrees in increments of $5$ degrees.

\subsection{Myo Dataset}
Dataset (G) was acquired using Myo bracelet~\cite{Myo} that provides orientation information of the bracelet as a quaternion data stream.
We selected $5$ popular gestures from the American Sign Language: \textit{hello}, \textit{good bye}, \textit{thank you}, \textit{you're welcome}, \textit{please}.
We captured $4$ subjects performing all these gestures $10$ times for a total of $200$ instances.

\subsection{Lip Motion Dataset}
We captured dataset (H) consisting of face data for $22$ participants uttering the password {\emph siggraph rocks}. 
We recorded the uttering using a video camera, we ran a face tracker~\cite{saragih2009face} and we saved 
the 2D positions of the $16$ points around the mouth.

\subsection{Kickboxing Dataset}
We created a video database (I) for kickboxing activity as part of project to learn and embed an individual actor\textsc{\char39}s action sequence style in 3D game characters. This database contains 270 video sequences, consisting of seven basic kickboxing actions - Jab, HooK, Uppercut (UC), Defense (Def), Side Kick (SK), Lower Kick (LK), and Jumping Defense (JD) recorded for 10 actors from 3 different views (front view and two side views). For LK and SK we have recorded from six views. All the videos are taken with a stationary camera; however, backgrounds as well as actors\textsc{\char39} clothes/positions/orientations may differ. The videos are down-sampled to spatial resolution of 160X120 pixels and a frame rate of 20fps. These videos have an average length of 15 seconds.



\end{document}